%% file: article.tex
\documentclass[pmlr]{jmlr}%

\usepackage{longtable}%

\usepackage{booktabs}

\usepackage{multirow}
\usepackage{graphicx}
\usepackage{color}
\usepackage{natbib}
\usepackage{dsfont}
\usepackage{amsmath}
\usepackage{comment}
\usepackage{centernot}
\usepackage{paralist}
\usepackage{hyperref}
\usepackage{tikz}
\usepackage{tkz-graph}
\usetikzlibrary{shapes.geometric}
\usetikzlibrary{backgrounds}
\usetikzlibrary{arrows}
\usetikzlibrary{patterns}

\newcommand{\transpose}{\mbox{${}^{\text{T}}$}}

\definecolor{linkcolor}{HTML}{023C40}
\definecolor{citecolor}{HTML}{1F5673}
\definecolor{urlcolor}{HTML}{A52422}

\hypersetup{
  linkcolor  = linkcolor,
  citecolor  = citecolor,
  urlcolor   = urlcolor,
  colorlinks = true,
}

\usepackage[load-configurations=version-1]{siunitx} %
\theorembodyfont{\upshape}
\theoremheaderfont{\scshape}
\theorempostheader{:}
\theoremsep{\newline}

\title[Learning by Doing:
Controlling a Dyn.\  System using
Causality, Control, and
RL]{
Learning by Doing:
Controlling a Dynamical System using
Causality, Control, and
Reinforcement Learning
}

\author{
\Name{Sebastian Weichwald} \Email{sweichwald@math.ku.dk}\\
\addr Department of Mathematical Sciences, University of Copenhagen, Denmark
\AND
\Name{Søren Wengel Mogensen} \Email{soren.wengel\_mogensen@control.lth.se}\\
\addr Department of Automatic Control, Lund University, Sweden
\AND
\Name{Tabitha Edith Lee} \Email{tabithalee@cmu.edu}\\
\addr The Robotics Institute, Carnegie Mellon University, USA
\AND
\Name{Dominik Baumann} \Email{dominik.baumann@it.uu.se}\\
\addr Department of Information Technology, Uppsala University, Sweden
\AND
\Name{Oliver Kroemer} \Email{okroemer@cmu.edu}\\
\addr The Robotics Institute, Carnegie Mellon University, USA
\AND
\Name{Isabelle Guyon} \Email{guyon@chalearn.org}\\
\addr LISN/INRIA/CNRS, Université Paris-Saclay, France, and ChaLearn, USA
\AND
\Name{Sebastian Trimpe} \Email{trimpe@dsme.rwth-aachen.de}\\
\addr Institute for Data Science in Mechanical Engineering, RWTH Aachen University, Germany
\AND
\Name{Jonas Peters} \Email{jonas.peters@math.ku.dk}\\
\Name{Niklas Pfister} \Email{np@math.ku.dk}\\
\addr Department of Mathematical Sciences, University of Copenhagen, Denmark
}

\begin{document}

\maketitle

\begin{abstract}
Questions in causality, control, and reinforcement learning  go beyond the classical machine learning task of prediction under i.i.d.\ observations.
Instead, these fields consider the problem of learning how to actively perturb a
system to achieve a certain effect on a response variable.
Arguably, they have complementary views on the problem:
In control,
one usually aims to first identify the system by excitation strategies to then apply model-based design techniques to control the system.
In (non-model-based) reinforcement learning, one directly optimizes a reward.
In causality, one focus is on identifiability of causal structure.
We believe that combining the different views might create synergies and this competition is meant as a first step toward such synergies.
The participants had access to
observational and  (offline) interventional data generated by
dynamical systems.
Track~CHEM considers an open-loop problem in which a single impulse at the beginning of the dynamics can be set, while Track~ROBO considers a closed-loop problem in which control variables can be set at each time step.
The goal in both tracks is to infer controls
that drive the system to a desired state.
Code is open-sourced (\href{https://github.com/LearningByDoingCompetition/learningbydoing-comp}{github.com/LearningByDoingCompetition/learningbydoing-comp})
to reproduce the winning solutions of the competition and
to facilitate trying out new methods on the competition tasks.
\end{abstract}

\section{Introduction}

Modeling {actively performed} changes in an observed system is an important goal that has appeared in various versions and settings in statistics, engineering, and computer science.
Each community has developed their own terminology and methods to tackle the specific applications relevant to their disciplines, leading to the emergence of causality, control theory, and reinforcement learning (RL).
Each of these fields brings a different perspective to modeling {system changes} and our goal of the \emph{Learning by Doing} NeurIPS 2021 competition was to bring together researchers from each of these fields to work on the same set of tasks.

We decided to focus on dynamical systems as these appear in all three fields. To offer a sufficiently diverse set of problems, we provided two competition tracks: Track~CHEM and Track~ROBO. Track~CHEM considers the open-loop problem of choosing a single impulse that can be set at the beginning of a chemical reaction with the goal of reaching a specific concentration of a target reactant. Track~ROBO considers the closed-loop problem of continuously providing inputs to a robot that guides the tip of the robot to move along a target trajectory. In both cases, participants were given a recorded data set from the systems and needed to use this data to learn how to optimally interact with the system in new settings.

In Section~\ref{sec:threefields}, we
{briefly}
introduce the three fields, provide
the relevant terminology, and discuss how each field models exogenous changes to a system.
In Section~\ref{ssec:track1OptControl}, we introduce Track~CHEM and in Section~\ref{sec:ROBO} Track~ROBO.
In both aforementioned sections, we also point out the challenges the tasks pose and how they relate to each field.
We conclude in Section~\ref{sec:conclusion} with some of the lessons learnt throughout the competition.

The competition website
\href{https://learningbydoingcompetition.github.io/}{learningbydoingcompetition.github.io}
provides tutorials, results, and presentations of some of the competing teams.
We provide open-source code
at \href{https://github.com/LearningByDoingCompetition/learningbydoing-comp} {github.com/LearningByDoingCompetition/learningbydoing-comp}
to reproduce the winning solutions and
to allow the application of new methods to the competition tasks.

\section{Causality, Control, and Reinforcement Learning}\label{sec:threefields}

To foster cross-pollination, we begin by introducing causality, control, and RL and describe how each framework models changes to a system.

\paragraph{Causality}

In classical statistics, a multivariate stochastic system is thought of as describing a single observational distribution. In contrast, a causal system\footnote{%
Here, the notion of a causal system
differs from what is usually called a causal system in the systems and control literature
where it refers to systems in which outputs only depend on past and current inputs.}
describes a set of distributions that models system behavior not only under passive observation, but also under interventions \citep{Pearl2009, Spirtes2000, Imbens2015}.
To make this more precise, assume we observe a response variable $Y$ and a set of predictors $X=(X_1,\ldots,X_p)$ and wish to model how $Y$ is affected by interventions on the predictors $X$. We now assume that there exists a subset $\text{PA}\subseteq\{1,\ldots,p\}$ of the predictors, called the parents of $Y$, that determine the value of the response $Y$ via the following fixed functional form
\begin{equation}
    \label{eq:causal_eq}
    Y=f(X_{\text{PA}},\epsilon),
\end{equation}
where $\epsilon$ is a noise variable. This equation is understood to be functional in the sense that the expected value of $Y$ given that $X_{\text{PA}}$ was set to a fixed value $x_{\text{PA}}$ (denoted by $\mathbb{E}[Y\mid \text{do}(X=x_\text{PA})]$) is given by $\mathbb{E}[f(x_{\text{PA}},\epsilon)]$.
Such structural equations are the building blocks of structural causal models (SCMs) \citep{Pearl2009, Peters2017book}, which is an important class of causal models. Usually, causal models assume some version of stability of mechanisms under interventions  \citep{Haavelmo1944, Aldrich1989}. In \eqref{eq:causal_eq} this corresponds to assuming that $f$ remains fixed under any intervention on $X$. Much research in causality investigates under which conditions the function $f$ and the set of parent variables $X_{\text{PA}}$ can be identified from data {and how to do so data-efficiently}.
Causal models also exist for dynamical systems, for example, a %
{multivariate process} $Z(t)=(Z_1(t),\ldots,Z_p(t))$ can be modeled by differential equations of the form $dZ_j(t) = F(Z(t))_j$ for each {component}. Interventions then correspond to modifying parts of these equations, for example, by fixing one of the coordinate processes at a certain value over a given time interval \citep[see][for an overview]{Peters2020dynamics}.
Causal models induce a \emph{graph} over the coordinate processes{, which} is useful for visualizing a multivariate causal system. In such a graph, each node represents a coordinate process, $Z_i$. We include a \emph{directed edge}, $Z_i \rightarrow Z_j$, $i\neq j$, if
the right-hand side of $dZ_j(t) = F(Z(t))_j$ is not constant in $Z_i$
and in this case we say that $Z_i$ is a \emph{(causal) parent} of $Z_j$ and that $Z_j$ is a \emph{child} of $Z_i$.

\paragraph{Control} In automatic control, one assumes that a target process $y:[0,T]\rightarrow\mathbb{R}^p$ is generated by a dynamical system. A common setup is a continuous-time state-space model, given by
\begin{align}
    \dot{x}(t) &= F(x(t), u(t))f \label{eq:control_equation} \\
    y(t) &= H(x(t),u(t)),\nonumber
\end{align}
where $x(t)$ is the time-dependent \emph{state} of the systems (which could include $y(t)$ itself), $\dot{x}(t)$ the derivative of $x$ with respect to time, and $u(t)$ is the \emph{control input}.
\emph{Control design} is the task of constructing a \emph{controller}, that is, a map from measurements $y(t)$ to control inputs $u(t)$.
The dynamical system~\eqref{eq:control_equation}, sometimes  called a \emph{plant}, is often nonlinear in the state $x(t)$, and linear in the control input $u(t)$. When both the inputs $x(t)$ and $u(t)$ and the target process are  vector-valued, one also uses the term \emph{multi-input multi-output} (MIMO) system.

A typical approach of control engineering for obtaining a controller may consist of the following steps:
(1) Use problem insight to select a useful (often parametric) model class (such as linear/nonlinear, auto-regressive, continuous/discrete-time);
(2) Fit the parameters of the model (that have not been set yet);
(3) Use the model in a control design method to obtain a controller (for example, by minimizing a given loss function); (4) Test the controller on the system or in simulation;
(5) Possibly repeat the cycle if results are not satisfactory.

Control design
traditionally builds model classes from first principles, for example,
laws in chemistry or physics, even though not all parameters may be known.
The task of fitting the model from input-output data is known as \emph{system identification} \citep[see, for example,][]{ljung1998system}.
Identifying a system and obtaining a controller is a well-understood problem for linear dynamical systems
and we refer the reader to textbooks such as the one by \citet{astromMurray} for a more detailed introduction into system identification and control design.

For nonlinear systems, however, many design methods exist for specific problem settings, but they lack the generality of approaches for linear systems.
Furthermore, in some practical applications,
including the setup of this competition,
a derivation from first order principles (including all parameters) may be impossible and even the model class may be unknown.
In slight deviation of items (1--3) above,
one can attempt to control the system without an explicit model of the underlying dynamics, for example, using a PID controller \citep{astromMurray}.
Alternatively, \emph{model-predictive control}  \citep{allgower2012nonlinear, borrelli2017predictive}
and similar optimization-based controller schemes may not compute a controller explicitly but model
the effect of control inputs directly and exploit this model in an online optimization procedure.
While model-predictive control typically relies on a given system model with fixed parameters, the field of \emph{adaptive control} \citep{aastrom2013adaptive} considers settings where model or controller parameters need to be tuned online.
More recently, researchers in control have been exploring ways to incorporate data-based and machine-learning approaches into control design, partly making the above pipeline more flexible. This emerging area between control and machine learning is known as \emph{learning-based control}
or
\emph{data-based control}.

\paragraph{Reinforcement learning}
In RL, one commonly starts by defining a \emph{reward} that specifies how desirable it is to transition from one state to another under a given action. %
The system is modeled sequentially by explicitly accounting for interactions with the system in each time step. The goal is to learn how to interact with the system in a way that maximizes the expected value of the reward. Mathematically, this {can be} achieved using Markov Decision Processes (MDPs). An MDP consists of a tuple $\left(S,A,R,T,\gamma \right)$ \citep{Sutton1998}. $S$ is the set of states of the system and $A$ is a set of actions that the agent can execute.
The reward $R(s,a,s')$ expresses the immediate reward for executing action $a\in A$ in state $s\in S$ and then transitioning to the next state $s' \in S$. $T(s'| s,a)$ is the transition distribution which gives the distribution over next states $s'$ given the current state $s$ and  action $a$. $\gamma \in \left[0,1\right]$ is a  discount factor that expresses the agent's preference for immediate rewards over long-term  rewards.
{Usually, $S$, $A$ and $\gamma$ are known (user-defined), $T$ and $R$ are unknown.}
To select an action,  the agent applies a policy $\pi(a|s)$ that defines the distribution over the next action, $a$, to execute given the current state $s$. Policies can be stochastic or deterministic. The $t$-th sampled transition thus results in a tuple $(s_t,a_t,s_t',r_t)$, where $s_t$ is the current state, $a_t$ is the sampled action, $s_t'$ is the next state after the transition, and $r_t\in \mathbb{R}$ is the resulting scalar reward. The goal of learning is to acquire an optimal policy, often denoted as $\pi^*$, that maximizes the expected return $\mathbb{E}_{s'\sim T,a \sim \pi}\left[\sum_t^T \gamma^t r_t\right]$ where T is the duration of the task \citep{Bell1996}.

\section{Track~CHEM: Optimally controlling a chemical reaction}\label{ssec:track1OptControl}

Track~CHEM tackles the problem of optimally choosing impulses or shocks in a dynamical system to control a specific part of the system. This task is motivated by applications related to chemical reactions in which one is interested in generating a desired concentration of a specific chemical compound by controlling the initial concentration of some other chemical compounds. When considering such systems there are constraints on how and at what cost experimentation can be performed. To reflect this, we only provided participants with offline training data instead of allowing them to actively interact with the reactions.
The task was to extract knowledge from observed experiments, and use it to control the system in previously unseen settings. Methods that tackle this problem may also apply to systems in which experimentation is infeasible and instead only exogenous shocks to the system can be observed and leveraged for learning.

\paragraph{Background on chemical reaction networks}
In a chemical reaction, one set of
chemical compounds is transformed into another. We usually say that
reactants are turned into products. Reactants and products are both called species. In Track~CHEM, the goal is to find optimal controls (or policies or interventions) on the concentrations of reactants to ensure a desired concentration of one of the species. The dynamical behaviour of species concentrations in chemical reactions is modeled by mass-action kinetics \citep{Waage64}, which results in an ordinary differential equation (ODE) over the species.
During training, the participants were not able to interact with the system and instead only had access to past observations from the system.
For these observations, the applied control inputs were known to participants.
The goal in this track is to control one specific process in the observed system when provided only with
initial observations. %
We give more details on chemical reaction networks in Appendix~\ref{apd:chem}.

\paragraph{Data generating process}
Data is generated by an artificial chemical reaction network.
Specifically, a $15$-dimensional
process $Z(t)_{t \geq 0}$
is generated as:
\begin{equation}
    \begin{split}
         Z(0)&=z\\
        \dot{Z}(t)&=F(Z(t)) + B U(t),
    \end{split}
    \label{eq:dgp_task1}
\end{equation}
where $U(t)\in [-10,10]^8$ is the control input at time $t$, $z\in (0,\infty)^{15}$ is an initial value, $B\in\mathbb{R}^{15\times 8}$
is a matrix specifying how the controls influence the dynamics and $F:\mathbb{R}^{15}\rightarrow\mathbb{R}^{15}$ is a function from the function class
\begin{equation}
    \mathcal{F}=\left\{F:\mathbb{R}^{15}\rightarrow\mathbb{R}^{15}\,\vert\,  F_{\ell}(Z)=\sum_{j=1}^{15}\theta_j^{\ell} Z_j+\sum_{k,j=1}^{15}\theta_{j,k}^{\ell} Z_j Z_k,\ \ \ell = 1,\ldots,{15} \right\}.
    \label{eq:observationModelTask1}
\end{equation}
The parameter $\theta$ satisfies additional constraints since we only consider ODE systems that are generated by converting chemical reactions using the law of mass-action kinetics.
Furthermore, the rates of the underlying chemical reactions are non-negative which also adds constraints on the coefficients $\theta$.

Among the $15$ species, $Y = Z_{15}$ is the
species for which the concentration should be controlled. There are eight controls, $U_1(t),\ldots, U_8(t)$, that affect the concentrations of a subset of the species.
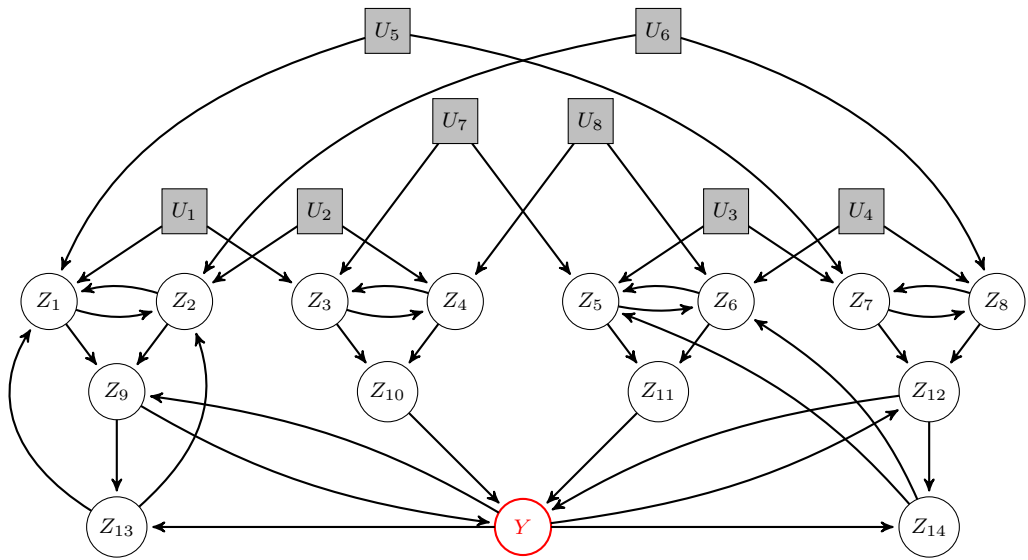
\begin{figure}
    \centering
    \scriptsize
    \begin{tikzpicture}[xscale=0.9, yscale = 0.6]
    \tikzstyle{VertexStyle} = [shape = circle, minimum width = 2.5em]
      \tikzstyle{VertexStyle} = [shape = circle, minimum width = 2.5em,draw]
      \SetGraphUnit{2}
      \Vertex[Math,L=Z_1,x=-7,y=4]{X1}
     \Vertex[Math,L=Z_2,x=-5,y=4]{X2}
      \Vertex[Math,L=Z_3,x=-3,y=4]{X3}
    \Vertex[Math,L=Z_4,x=-1,y=4]{X4}
      \Vertex[Math,L=Z_5,x=1,y=4]{X5}
  \Vertex[Math,L=Z_6,x=3,y=4]{X6}
  \Vertex[Math,L=Z_7,x=5,y=4]{X7}
  \Vertex[Math,L=Z_8,x=7,y=4]{X8}
  \Vertex[Math,L=Z_9,x=-6,y=2]{X9}
  \Vertex[Math,L=Z_{10},x=-2,y=2]{X10}
  \Vertex[Math,L=Z_{11},x=2,y=2]{X11}
  \Vertex[Math,L=Z_{12},x=6,y=2]{X12}
  \Vertex[Math,L=Z_{13},x=-6,y=-1]{X13}
  \Vertex[Math,L=Z_{14},x=6,y=-1]{X14}
  \tikzstyle{VertexStyle} = [shape = circle, minimum width =
  2.5em, draw, color=red, thick]
  \Vertex[Math,L=Y,x=0,y=-1]{Y}
  \tikzstyle{VertexStyle} = [shape = rectangle, minimum width = 2.0em,
  minimum height = 2.0em, draw, fill=lightgray]
  \Vertex[Math,L=U_1,x=-5,y=6]{U1}
  \Vertex[Math,L=U_2,x=-3,y=6]{U2}
  \Vertex[Math,L=U_3,x=3,y=6]{U3}
  \Vertex[Math,L=U_4,x=5,y=6]{U4}
  \Vertex[Math,L=U_5,x=-2,y=10]{U5}
  \Vertex[Math,L=U_6,x=2,y=10]{U6}
  \Vertex[Math,L=U_7,x=-1,y=8]{U7}
  \Vertex[Math,L=U_8,x=1,y=8]{U8}
  \tikzstyle{EdgeStyle} = [->,>=stealth',shorten > = 2pt]
  \Edge(X1)(X9)
  \Edge(X2)(X9)
  \Edge(X3)(X10)
  \Edge(X4)(X10)
  \Edge(X5)(X11)
  \Edge(X6)(X11)
  \Edge(X7)(X12)
  \Edge(X8)(X12)
  \Edge(X9)(X13)
  \Edge(Y)(X13)
  \Edge(X12)(X14)
  \Edge(Y)(X14)
  \Edge(X10)(Y)
  \Edge(X11)(Y)
  \Edge(U1)(X1)
  \Edge(U1)(X3)
  \Edge(U2)(X2)
  \Edge(U2)(X4)
  \Edge(U3)(X5)
  \Edge(U3)(X7)
  \Edge(U4)(X6)
  \Edge(U4)(X8)
  \Edge(U7)(X3)
  \Edge(U8)(X4)
  \Edge(U7)(X5)
  \Edge(U8)(X6)
  \tikzset{EdgeStyle/.append style = {->, bend right=25}}
  \Edge(X2)(X1)
  \Edge(X1)(X2)
  \Edge(X4)(X3)
  \Edge(X3)(X4)
  \Edge(X6)(X5)
  \Edge(X8)(X7)
  \Edge(X7)(X8)
  \Edge(U5)(X1)
  \Edge(U6)(X2)
  \Edge(X13)(X2)
  \tikzset{EdgeStyle/.append style = {->, bend left=25}}
  \Edge(U5)(X7)
  \Edge(U6)(X8)
  \tikzset{EdgeStyle/.append style = {->, bend left=35}}
  \Edge(X13)(X1)
  \tikzset{EdgeStyle/.append style = {->, bend right=15}}
  \Edge(X9)(Y)
  \Edge(Y)(X9)
  \Edge(X12)(Y)
  \Edge(Y)(X12)
  \Edge(X14)(X5)
  \Edge(X14)(X6)
  \Edge(X5)(X6)
\end{tikzpicture}
\caption{Graphical representation of the chemical reaction in Track~CHEM.
}
    \label{fig:my_label}
\end{figure}
A (to participants unknown) graphical representation of the model {that generated the data for the competition} is given in Figure~\ref{fig:my_label}. The model has a simple structure consisting of four blocks of variables:
$\{Z_1,Z_2,Z_9,Z_{13}\}$,
$\{Z_3,Z_4,Z_{10}\}$,
$\{Z_5,Z_6,Z_{11}\}$,
and
$\{Z_7,Z_8,Z_{12},Z_{14}\}$.
Each block corresponds to an interaction mechanism that can either increase or decrease the concentration of $Y$. There are two types of controls. (1) Control variables $U_1,U_2,U_3$, and $U_4$ affect the system strongly and they affect an increasing and a decreasing block simultaneously. (2) Variables $U_5,U_6,U_7$, and $U_8$ %
have a weaker effect on the system but they target only an increasing block ($U_7$ and $U_8$) or only a decreasing block ($U_5$ and $U_6$).
Hence, the controls in (2) offer an easy strategy to control $Y$ but are expensive, while using the controls in (1) is cheaper but might be more difficult.
The participants only observed data with pre-specified control settings
where we {had} incorporated confounding structure in the observed controls;
{the problem therefore also featured a causal challenge.}

\paragraph{Task}
Participants knew the function class but did not know the parameters. They did not know the graphical structure of the system{, either}. Furthermore, participants did not observe the process $Z$ directly. Instead, they only observed $X$, a noisy version of the process, sampled on a time grid $(t_0,\ldots,t_L)$;
that is,
{the observed data is sampled,
at $t \in \{t_0, \ldots, t_L\}$,
from the process}
\begin{equation} \label{eq:observdnoise}
    X(t)=
(Z_1(t),\ldots, Z_{15}(t)) + N(t),
\end{equation}
where $N$ is a mean-zero noise process such that $\{N(t): t = t_0,\ldots,t_L\}$ are independent.
The goal of the task is to choose a value $u\in\mathbb{R}^8$ such that the controls
\begin{equation}
    U(t)=
    \begin{cases}
    u \quad \text{if } t \in [t_0,t_3)\\
    0 \quad \text{if } t \geq t_3,
    \end{cases}
    \label{eq:controlImpulseTask1}
\end{equation}
lead to $Y$
being close to a (pre-specified) desired value $y_*$ at the end of the reaction.
As the value $u$ for the control is set only once {and} after having observed $X(0)$,
{Track~CHEM is an open-loop problem.}

Participants had access to data from $12$ different ODE systems which are specified by the functions $F^1,\ldots,F^{12}\in\mathcal{F}$ and they knew the index of the system from which data originated. The function class $\mathcal{F}$ was known but $F^1,\ldots,F^{12}$ were unknown to the participants. Participants knew that the $12$ systems had the same structure, that is, the same parameters $\theta_j^{\ell}$ and $\theta_{j,k}^{\ell}$ were zero in all $12$ systems, and that every $\theta_j^{\ell}$ and $\theta_{j,k}^{\ell}$ had the same sign in all systems. The parameters of the noise as well as the matrix $B$ were the same in all $12$ systems and these facts were also known to participants.

The training data available to participants was
generated by running the data generating process $20$ times for each $F^i$ with different pairs of initial conditions $z$ and controls $u$. The distributions used to select $z$ and $u$ in the training data were unknown to participants and differed among systems.

\paragraph{Evaluation}
For each of the systems participants were provided with $50$ additional sets of initial vectors, $X(0)$, as well as an indicator specifying the corresponding system ($i = 1,\ldots,12$). For each of these combinations participants were asked to select a control input to minimize the loss function. The loss function measures the proximity of $Y^{i,k}$ to the desired value $y^{i,k}_*$ toward the end of the observation interval while also adding a penalty term depending on the size of the control input used. The exact loss function can be found in Appendix \ref{apd:chem}.

\paragraph{Three perspectives}

\emph{Causality} The system is causal in the sense that it specifies not only an observational distribution ($U \equiv 0$ in Equation (\ref{eq:dgp_task1})) but also a set of interventional distributions ($BU \centernot\equiv 0$
in Equation (\ref{eq:dgp_task1})). The intrinsic dynamics are the same regardless of which (if any) intervention is applied as they are described by the function $F$. That is, the mechanism described by {$F$} is stable under interventions. The task can be thought of as a causal learning task where participants should predict the effect of interventions and choose an optimal intervention.

\emph{Control theory} The task seeks a functional map from measurements $y(t)$ to control inputs $u(t)$ which is the classical task of \emph{control design}.  The control inputs are to be of the form \eqref{eq:controlImpulseTask1}, which can be understood as \emph{impulse control}.  Formalizing the control objective as an optimization problem is commonly known as \emph{optimal control} (see, for example, \cite{bertsekas2000dynamic,anderson2007optimal}).
The objective function is a weighted sum that, as typical in control applications, balances \emph{control performance} and \emph{control effort}.

\emph{Reinforcement learning} In this task, the  vector $u$
is selected at the start of each trial and then executed for a number of steps.
This problem formulation without state transitions
is closely connected to bandit or contextual bandit problems \citep{Sutton1998}
where
the agent receives a reward based solely on the selected action and context. Classical online RL approaches learn the policy by iteratively interacting with the environment and improving the policy. However, for the proposed task, the agent will need to use \emph{offline} reinforcement learning as no interaction with the system is possible \citep[see, for example,][]{Lange2012,Levine2020ORL}.

\section{Track~ROBO: Controlling a robotic arm in a dynamical environment}\label{sec:ROBO}

Track~ROBO is motivated by the
long-term goal
of learning skills that can be performed by a diverse set of robots to complete real-world tasks.
For instance, one may want to teach robots new skills such as stirring a pot or cutting vegetables for cooking.
The new skills can be learned more efficiently by leveraging prior experience in related tasks such as whisking eggs.
Robots may also have different kinematic structures, requiring individualized control policies to accurately execute the end-effector trajectory required by a new skill.
Even robots of the same type can differ because of minor variations in the production process.
If we can leverage a robot's prior movement data to derive an individualized controller for a new skill,
we may avoid the need for additional training and enable rapid roll-out of new skills.

We mimic this challenge in Track~ROBO:
participants are provided with movement data and asked to provide a controller that sequentially interacts with a robotic arm such that its end-effector reaches a target position provided for the next time step.
The two difficulties are:
(1) Participants can only set abstract control variables
instead of, for example, setting the torques of individual joints directly.
This restriction imitates a setting in which the robot dynamics are complicated to write down explicitly.
(2) The training data is comprised of different types of trajectories than those in the test data, imitating a setting in which the robot must adjust to a new task given previous data of an old task.

\begin{figure}
    \centering
    \begin{minipage}{0.5\textwidth}
    \centering
    \includegraphics[width=1.1\textwidth]{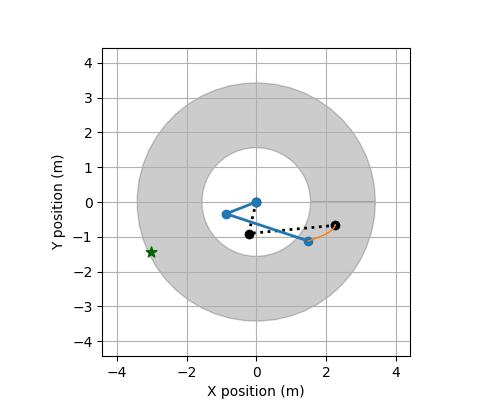}
    \end{minipage}%
    \begin{minipage}{0.5\textwidth}
    \centering
    \includegraphics[width=1.1\textwidth]{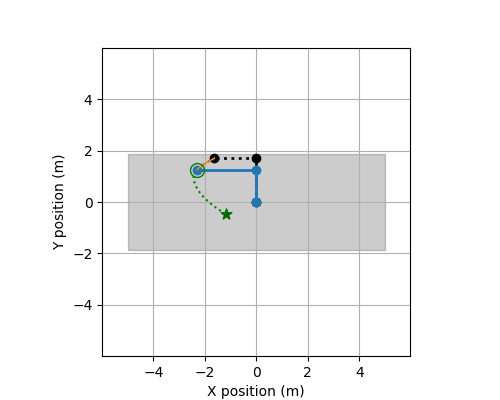}
    \end{minipage}
    \caption{(Left) Rotational 2-link robot{, the trail of previous positions of the robot tip (orange line), and the} target position (green star). (Right) Prismatic 2-link robot {and} a target trajectory (green dotted line) to a target position (green star). The gray area corresponds to the reachable workspace of the robots. The black dotted lines indicate the initial position of the robots.}
    \label{fig:example_robots}
\end{figure}

We consider three robot arms:
(1) A two-joint rotational robot arm,
(2) a three-joint rotational robot arm,
and (3) a two-joint prismatic robot arm.
The rotational joints produce a rotary motion around the joint, and the prismatic joints produce a linear motion between links (see Figure~\ref{fig:example_robots}).
Each joint can be controlled by applying a voltage signal to a DC motor located in the joint.
{(In this challenge, the voltage is not set directly, see below.)}
In rotational joints, this creates a torque, while in prismatic joints this creates a linear acceleration.
The resulting movement of the joints of the robot arm (and its tip in particular) are governed by the physical laws of motion.
Given a specific robot one can derive exact differential equations that describe the robot's
movement,
known as the \emph{dynamics model}
(or \emph{dynamics} for short) of the robot with parameters that depend on various specifications of the robot such as link mass, rotational moment of inertia, link length, location of center of mass, and friction coefficients, see Appendix~\ref{app:robots}.

\paragraph{Data generating processes}
Participants were able to sequentially interact with $24$ different robotic arm systems,
with dynamics given by
\begin{equation}
    \begin{split}
        (Z(0), W(0))&=(z, w)\\
        (\dot{Z}(t),\dot{W}(t))&=F^s(Z(t), W(t), C(t)) = F^s(Z(t), W(t), A^s\cdot U(t)),
    \end{split}
    \label{eq:dgp_task2}
\end{equation}
for $s\in\{1,...,24\}$
where $Z(t)=((X(t),Y(t))\in\mathbb{R}^{2}$ is the position of the tip of the robot,
the positions of other joints of the robot are
$W(t)=(X_1(t),Y_1(t), ...,X_d(t),Y_d(t))\in\mathbb{R}^{2d}$,
$z\in\mathbb{R}^{2}$ and $w\in\mathbb{R}^{2d}$ are the initial values,
and $C(t)\in\mathbb{R}^q$ are the underlying robot controls, that is, voltage signals applied to DC motors located in each joint
($d$ and $q$ depend on the underlying robot).
The (to participants unknown) functions $F^s:\mathbb{R}^{2(d+1)+q}\rightarrow\mathbb{R}^{2(d+1)}$ are given by the second-order dynamic system of the underlying robots:
$F^1$, $...$, $F^8$ and $F^9$, $...$, $F^{16}$ correspond to 2-link ($d=1$) and 3-link ($d=2$) open chain planar manipulators with revolute joints (cf.\ \ref{app:robots-rotational}), respectively,
and $F^{17}$, $...$, $F^{24}$ correspond to 2-link ($d=1$) prismatic manipulators (cf.\ \ref{app:robots-prismatic});
the dynamics differ further between robots due to different robot specifications denoted by $\theta^s$ (link masses and lengths, moments of inertia, friction coefficients, and locations of link center of masses).
The (to participants unknown) \emph{interface function}
$G^s:\mathbb{R}^p\rightarrow\mathbb{R}^q, x\mapsto A^s x$
for $A^s \in \mathbb{R}^{q\times p}$
relates the participants' abstract control inputs $U(t)\in\mathbb{R}^p$
to the underlying robots' control inputs $C(t)$ via $C(t)=G^s(U(t))$;
$A^s$ is either the identity, a square ($p=q$), or
a rectangular ($p > q$) real matrix with imbalanced row-norms and full rank.
Participants can control the systems only on a linearly spaced discrete time grid
$(t_0, t_1,\ldots, t_{200})$ with $t_0=0$ and $t_{200}=2$, that is,
for each time step $\ell\in\{0,\ldots,199\}$ it holds that
$C(t)=G^s(U(t))\equiv\text{const}$ for all $t\in[t_{\ell}, t_{\ell+1})$.
Some of the $24$ systems share robot dynamics ($F^s$), specifications ($\theta^s$), and/or the control interfaces ($A^s$) in a systematic way,
which is reflected in the naming convention but was not explicitly revealed to the participants
(cf.\ Table~\ref{tab:robots} for an overview).

\paragraph{Task}
The competition task is to control the robots' end-effector position $Z(t)$
to follow a given target process ${t \mapsto} z_*(t)$.
More specifically, participants needed to implement a controller, that is, a function
$\mathtt{controller}^s: \mathbb{R}^{2(d+1)+2}\to \mathbb{R}^{p}$
for each robot ($s\in\{1,...,24\}$).
At each time step $\ell\in\{0,\ldots,199\}$,
the controller is queried for the next control input $U(t_\ell)\in\mathbb{R}^p$
given
the current positions $Z(t_{\ell}), W(t_{\ell}) \in\mathbb{R}^{d+1}$,
their derivatives $\dot{Z}(t_{\ell}), \dot{W}(t_{\ell})\in\mathbb{R}^{d+1}$,
and a target end-effector position $z_*(t_{\ell+1})\in\mathbb{R}^2$ for the next time step.
The task does not involve planning as the controller only gets access to the next time step's end-effector target position, however,
the implemented controller can gather information during the control process\footnote{%
Even though $\mathtt{controller}^s$ is specified as a function of the current state and the next target state of the system,
participants were able to log previous queries allowing them to use the entire
past
$Z(t_{0}), W(t_{0}), ..., Z(t_{\ell}), W(t_{\ell})$,
and $\dot{Z}(t_{0}), \dot{W}(t_{0}), ..., \dot{Z}(t_{\ell}), \dot{W}(t_{\ell})$.}.
If the controller does not return within given compute time and resource constraints,
we set $C(t_\ell) = G^s(U(t_\ell)) = \mathbf{0}$.
This way, the different robots are propagated forward for different target trajectories
$(z_*(t_0),...,z_*(t_{200}))$ following their respective dynamics under the participant-provided controller;
the participants' task is to align the resulting end-effector trajectory $(Z(t_0),...,Z(t_{200}))$ with the target trajectory.

For deriving and implementing their controllers, participants were provided with (offline) training data
for each system $(F^s, \theta^s, A^s)$
in the form of $50$ realized end-effector trajectories and corresponding control input sequences.
Training trajectories are obtained using
an LQR-controller~\citep{anderson2007optimal}
(based on inverse dynamics and the (pseudo-)inverse of $G^s$ to map robot controls to participant controls)
to transition from some random starting positions to some random target positions in the robot's workspace.
The one-step ahead end-effector target positions used to generate the training trajectories were not provided to participants.
For each of these repetitions, participants were provided with the observed processes $W$ and $Z$, their derivatives $\dot{W}$ and $\dot{Z}$, a time indicator $t$, the applied controls $U$ and an indicator $i$ specifying the system.

\paragraph{Evaluation}
For each of the $24$ systems $(F^{1}, \theta^1, A^{1}),\ldots,(F^{24}, \theta^{24}, A^{24})$ the participants'
controller implementation is used to follow $10$ different target processes.
More specifically, for each system $i\in\{1,\ldots,24\}$ and repetition $k\in\{1,\ldots,10\}$, there is a target process $z_*^{i,k}:[0,2]\rightarrow \mathbb{R}^2$
and the robot is propagated forward using the participant-provided controller.
The loss function measures how far the realized end-effector trajectory is from the target process
and penalizes the size of the participants' control inputs
(cf.\ Appendix~\ref{app:robots} for details).
Mimicking a real-time robot control scenario, the participants' computational resources spent on evaluating the code that implemented their controller were restricted.
If the time constraints were not met, the submission was invalid.

\paragraph{Three perspectives}

\emph{Causality}
We can formulate the task of controlling a robot as a causal task. First, we need to estimate a model that allows us to evaluate the effect of various interventions, where interventions now correspond to setting the inputs $U(t)$. Second, we optimize a sequential intervention scheme.

\emph{Control}
The task consists of both system identification and controller design and it relies on all the steps in a typical control engineering application as outlined in Section \ref{sec:threefields}.

\emph{Reinforcement learning}  Track~ROBO has a standard MDP formulation wherein the action is defined by the command sent to the abstract controller and  the state space is given by the joint positions and velocities. The time-varying reward is given by the robot's accuracy in following the desired trajectory with a penalty on the size of the control inputs.

\section{Results and lessons learned}\label{sec:conclusion}
The results of the competition
can be found on our website\footnote{See
\href{https://learningbydoingcompetition.github.io/}{learningbydoingcompetition.github.io}.} and in the appendices.
The website also contains videos, in which some of the competing
teams describe their solutions in more detail.
Code is available, too\footnote{See \href{https://github.com/LearningByDoingCompetition/learningbydoing-comp} {github.com/LearningByDoingCompetition/learningbydoing-comp}
for open-source code
to reproduce the winning solutions of the competition and
to try out new methods on the competition tasks;
see \cite{joao_bravo_2022_5895099}
for code that implements the winning solutions to Track~CHEM and Track~ROBO;
see \href{https://github.com/Quarticai/learning_by_doing_solution}{github.com/Quarticai/learning\_by\_doing\_solution}
for code that implements the second winning solutions to Track~CHEM and Track~ROBO;
see \cite{Bussmann_A_Neural_Network_2022}
for code that implements the third winning solution to Track~CHEM;
see \cite{carlos_miguel_patino_2022_5888574}
for code that implements the third winning solution to Track~ROBO.
}.

For us, one of the key questions was whether, for the model classes considered in this challenge, it is better to aim to build a model and then infer the optimal control, or to directly estimate the effect of the applied control.
The competition results suggest the former in that in both tracks the winning solution was indeed inferring a data-generating model first (see in \ref{ssec:chemresults} and \ref{ssec:roboresults} in the appendices); it outperformed approaches of the latter type by a significant margin.
We speculate that imposing a model structure (even if both the structure and the parameters still need to be inferred from data) acts as strong regularization helping to ensure successful control which is robust to environmental changes.
Clearly, further research is needed to better understand in which settings this is expected to be the case.
In the future, it would be interesting to consider situations where the model inference becomes even harder, for example, because more parts of the system are unobserved.

\section*{Acknowledgments}
We thank the NeurIPS 2021 competition track organizers, Barbara Caputo, Marco Ciccone, and Douwe Kiela.
We also thank all the participants who took part in the competition.
SW and JP were supported by the Carlsberg Foundation.
SWM was supported by a DFF-International
Postdoctoral Grant (0164-00023B) from Independent Research Fund Denmark.
TL and OK were supported by the Office of Naval Research under Grant No.\ N00014-18-1-2775.
IG was supported by ANR Chair of Artificial Intelligence HUMANIA ANR-19-CHIA-0022.
JP was supported by a research grant (18968) from VILLUM FONDEN. NP was supported by a research grant (0069071) from Novo Nordisk Fonden.
The competition has been supported by the Department of Mathematical Sciences at the University of Copenhagen.

\clearpage
\bibliography{references}

\clearpage
\appendix

\section{More Details on Track CHEM}\label{apd:chem}

The following provides additional details on Track~CHEM of the competition.
\paragraph{Background on chemical reactions}
Parts of the following text are taken from
\citet{Peters2020dynamics}. A general reaction \citep[for example][]{Wilkinson09} takes the form
\begin{equation*}
  m_1 R_1 + m_2 R_2 + \ldots + m_r R_r \rightarrow
  n_1 P_1 + n_2 P_2 + \ldots + n_p P_p,
\end{equation*}
where $r$ is the number of reactants and $p$ the number of
products. Both $R_i$ and $P_j$ can be thought of as molecules and are
often called {species}. The coefficients $m_i$ and $n_j$ are positive integers, called
stoichiometries.

In mass-action kinetics \citep{Waage64}, one usually considers the
concentration $[X]$ of a species $X$, the square parentheses indicating
that one refers to the concentration
rather than to the
integer amount of a given species.
The concentration $[X]$ changes over time (but to simplify notation, we sometimes omit the notational dependence on $t$).
The law of mass-action allows one to convert the above equations into a system of ODEs over the concentrations of species. Formally, it states:
    \emph{The instantaneous rate of each reaction is proportional to
the product of each of its reactants raised to the power of its
stoichiometry.}
To better understand how this can be applied to transform reaction equations into a system of ODEs, it may help to consider an example. The Lotka-Volterra predator-pray model \citep{Lotka09} can be expressed in terms of reactions of the form
\begin{align}
  A &\overset{k_1}{\longrightarrow} 2A
      \label{eq:lotkareac1}\\
  A + B &\overset{k_2}{\longrightarrow} 2B
          \label{eq:lotkareac2}\\
  B & \overset{k_3}{\longrightarrow} \emptyset, \label{eq:lotkareac3}
\end{align}
where $A$ and $B$ describe abundance of prey and predators, respectively. In this model, the prey reproduce by themselves~\eqref{eq:lotkareac1},
but the predators require abundance of prey for reproduction, see~\eqref{eq:lotkareac2}.
After some time, also the predators die~\eqref{eq:lotkareac3}.
The coefficients
$k_1, k_2$, and $k_3$ indicate the rates, with which the reactions
happen
(the larger the rates, the faster the reactions).
Applying the law of mass-action
yields the following system of ordinary
differential equations (ODEs)
\begin{align}
  \tfrac{\text{d}}{\text{d}t}[A] &= k_1 [A] - k_2 [A][B] \label{eq:lotka1}\\
  \tfrac{\text{d}}{\text{d}t}[B] &= k_2 [A][B] - k_3 [B] \label{eq:lotka2}.
\end{align}

\paragraph{Chemical reactions of the data-generating process}

The data-generating process is illustrated in Figure \ref{fig:my_label}. The corresponding chemical reactions  are given by

\begin{align*}
  &Z_1 + Z_2 \overset{k_1}{\longrightarrow} Z_9
 \quad & Z_{10} \overset{k_5}{\longrightarrow} Y
 \quad\qquad & Z_{13} \overset{k_9}{\longrightarrow} Z_1 + Z_2\\
  &Z_3 + Z_4 \overset{k_2}{\longrightarrow} Z_{10}
  \quad &Z_{11} \overset{k_6}{\longrightarrow} Y
  \qquad\quad &Z_{14} \overset{k_{10}}{\longrightarrow} Z_5 + Z_6.\\
  &Z_5 + Z_6 \overset{k_3}{\longrightarrow} Z_{11}
  \quad &Z_9 + Y \overset{k_7}{\longrightarrow} Z_{13}
  \qquad\quad &\quad\\
  &Z_7 + Z_8 \overset{k_4}{\longrightarrow} Z_{12}
  \quad &Z_{12} + Y \overset{k_8}{\longrightarrow} Z_{14}
  \qquad\quad &\quad
\end{align*}

This system can be converted using the law of mass action resulting in the following ODE system.
\begin{equation*}
    \begin{split}
        &\tfrac{\text{d}}{\text{dt}}[Z_1] = -k_1[Z_1][Z_2] + k_9[Z_{13}]\\
        &\tfrac{\text{d}}{\text{dt}}[Z_2] = -k_1[Z_1][Z_2] + k_9[Z_{13}]\\
        &\tfrac{\text{d}}{\text{dt}}[Z_3] = -k_2[Z_3][Z_4]\\
        &\tfrac{\text{d}}{\text{dt}}[Z_4] = -k_2[Z_3][Z_4]\\
        &\tfrac{\text{d}}{\text{dt}}[Z_5] = -k_3[Z_5][Z_6] + k_{10}[Z_{14}]\\
        &\tfrac{\text{d}}{\text{dt}}[Z_6] = -k_3[Z_5][Z_6] + k_{10}[Z_{14}]\\
        &\tfrac{\text{d}}{\text{dt}}[Z_7] = -k_4[Z_7][Z_8]\\
        &\tfrac{\text{d}}{\text{dt}}[Z_8] = -k_4[Z_7][Z_8]
    \end{split}
    \qquad
    \begin{split}
        &\tfrac{\text{d}}{\text{dt}}[Z_9] = k_1[Z_1][Z_2]-k_7[Z_9][Y]\\
        &\tfrac{\text{d}}{\text{dt}}[Z_{10}] = k_2[Z_3][Z_4]-k_5[Z_{10}]\\
        &\tfrac{\text{d}}{\text{dt}}[Z_{11}] = k_3[Z_5][Z_6]-k_6[Z_{11}]\\
        &\tfrac{\text{d}}{\text{dt}}[Z_{12}] = k_4[Z_7][Z_8]-k_8[Z_{12}][Y]\\
        &\tfrac{\text{d}}{\text{dt}}[Z_{13}] = k_7[Z_9][Y]-k_9[Z_{13}]\\
        &\tfrac{\text{d}}{\text{dt}}[Z_{14}] = k_8[Z_{12}][Y]-k_{10}[Z_{14}]\\
        &\tfrac{\text{d}}{\text{dt}}[Y] = k_5[Z_{10}] + k_6[Z_{11}] - k_7[Z_9][Y] - k_8[Z_{12}][Y]
    \end{split}
\end{equation*}

\paragraph{Evaluation}

For each of the systems, $i = 1,\ldots,12$, partipants were asked to provide control input for $50$ initial values.
Participants' control inputs were evaluated by running the data-generating process for each of the provided controls to compute the following loss for each system\footnote{\label{sharedfootnote} The integrals are approximated numerically.}
\begin{equation}\label{eq:full_score_track1}
    J_i :=\frac{1}{50}\sum_{k=1}^{50}\left(\sqrt{\frac{1}{40}\int_{40}^{80}\big(Z_{15}^{i, k}(t)-y^{i, k}_*\big)^2 dt } + c\cdot \sqrt{\frac{||u^{i,k}||_2^2}{8}}\right),
\end{equation}
where $c=\frac{1}{20}$ and $u^{i,k} \in \mathbb{R}^p$ is the control input provided by the participant corresponding to the $k$th initial condition in the $i$th system. The process $Y^{i,k}$ of course depends on the provided input, $u^{i,k}$, even though this is not made explicit in the notation.

\subsection{CHEM results}
\label{ssec:chemresults}

The following table summarizes the results from Track CHEM. The keywords describing participants' solutions were chosen by the organizers based on participants' summaries of their solutions. \emph{Oracle} corresponds to a solution using access to the true data generating process. \emph{Oracle\textsuperscript{e}} corresponds to a solution generated with access to the true data generating process, but using only the expensive controls (see Section \ref{ssec:track1OptControl}). \emph{Zero} corresponds to a solution choosing $U \equiv 0$ for every system and initial condition.

\begin{center}
\begin{tabular}{ c | c | c | c | c}
Team name &  Score & Place & Keywords \\ \hline
\emph{Oracle} & 0.0872  &  & \\  \hline
 Ajoo &  0.0890 & 1st &  \begin{tabular}{c}Sparse estimation of graph \\ Direct estimation of a function in $\mathcal{F}$ \end{tabular}  \\ \hline
 \emph{Oracle\textsuperscript{e}} & 0.1450 &  & \\  \hline
  TeamQ &  0.3385 & 2nd & Neural network prediction of target  \\  \hline
   GuineaPig & 0.3386 &  3rd & Neural network prediction of target \\ \hline
   \emph{Zero} & 0.9686 &  &
\end{tabular}
\end{center}

\section{More Details on Track ROBO}\label{app:robots}

\begin{table}[]
    \centering
\begin{tabular}{c|c|c|c}
robot & dynamics ($F$) & specification ($\theta$) & interface ($A$) \\
\hline
great-devious-beetle &
\multirow{8}{*}{Rotational2} &
$\theta^{\text{gr-be}}$ &
$A^{\text{de}} = \mathbf{1}_{2 \times 2}$ \\
great-vivacious-beetle &
&
$\theta^{\text{gr-be}}$ &
$A^{\text{vi}}\in\mathbb{R}^{2 \times 2}$ \\
great-mauve-beetle &
&
$\theta^{\text{gr-be}}$ &
$A^{\text{ma}}\in\mathbb{R}^{2 \times 3}$ \\
great-wine-beetle &
&
$\theta^{\text{gr-be}}$ &
$A^{\text{wi}}\in\mathbb{R}^{2 \times 4}$ \\
rebel-devious-beetle &
&
$\theta^{\text{re-be}}$ &
$A^{\text{de}} = \mathbf{1}_{2 \times 2}$ \\
rebel-vivacious-beetle &
&
$\theta^{\text{re-be}}$ &
$A^{\text{vi}}\in\mathbb{R}^{2 \times 2}$ \\
rebel-mauve-beetle &
&
$\theta^{\text{re-be}}$ &
$A^{\text{ma}}\in\mathbb{R}^{2 \times 3}$ \\
rebel-wine-beetle &
&
$\theta^{\text{re-be}}$ &
$A^{\text{wi}}\in\mathbb{R}^{2 \times 4}$ \\
\hline
talented-ruddy-butterfly &
\multirow{8}{*}{Rotational3} &
$\theta^{\text{ta-bu}}$ &
$A^{\text{ru}}=\mathbf{1}_{3 \times 3}$ \\
talented-steel-butterfly &
&
$\theta^{\text{ta-bu}}$ &
$A^{\text{st}}\in\mathbb{R}^{3 \times 3}$ \\
talented-zippy-butterfly &
&
$\theta^{\text{ta-bu}}$ &
$A^{\text{zi}}\in\mathbb{R}^{3 \times 4}$ \\
talented-antique-butterfly &
&
$\theta^{\text{ta-bu}}$ &
$A^{\text{an}}\in\mathbb{R}^{3 \times 6}$ \\
thoughtful-ruddy-butterfly &
&
$\theta^{\text{th-bu}}$ &
$A^{\text{ru}}=\mathbf{1}_{3 \times 3}$ \\
thoughtful-steel-butterfly &
&
$\theta^{\text{th-bu}}$ &
$A^{\text{st}}\in\mathbb{R}^{3 \times 3}$ \\
thoughtful-zippy-butterfly &
&
$\theta^{\text{th-bu}}$ &
$A^{\text{zi}}\in\mathbb{R}^{3 \times 4}$ \\
thoughtful-antique-butterfly &
&
$\theta^{\text{th-bu}}$ &
$A^{\text{an}}\in\mathbb{R}^{3 \times 6}$ \\
\hline
great-piquant-bumblebee &
\multirow{8}{*}{Prismatic} &
$\theta^{\text{gr-bu}}$ &
$A^{\text{pi}}=\mathbf{1}_{2 \times 2}$ \\
great-bipedal-bumblebee &
&
$\theta^{\text{gr-bu}}$ &
$A^{\text{bi}}\in\mathbb{R}^{2 \times 2}$ \\
great-impartial-bumblebee &
&
$\theta^{\text{gr-bu}}$ &
$A^{\text{im}}\in\mathbb{R}^{2 \times 3}$ \\
great-proficient-bumblebee &
&
$\theta^{\text{gr-bu}}$ &
$A^{\text{pr}}\in\mathbb{R}^{2 \times 4}$ \\
lush-piquant-bumblebee &
&
$\theta^{\text{lu-bu}}$ &
$A^{\text{pi}}=\mathbf{1}_{2 \times 2}$ \\
lush-bipedal-bumblebee &
&
$\theta^{\text{lu-bu}}$ &
$A^{\text{bi}}\in\mathbb{R}^{2 \times 2}$ \\
lush-impartial-bumblebee &
&
$\theta^{\text{lu-bu}}$ &
$A^{\text{im}}\in\mathbb{R}^{2 \times 3}$ \\
lush-proficient-bumblebee &
&
$\theta^{\text{lu-bu}}$ &
$A^{\text{pr}}\in\mathbb{R}^{2 \times 4}$ \\
\end{tabular}
    \caption{Overview of the $24$ robot systems used in Track~ROBO.
    Here, $\theta^*$ refers to the robot specification (link lengths and masses, moments of inertia, friction coefficients, and locations of link center of masses)
    and $A^*\in\mathbb{R}^{q\times p}$ parametrizes the linear interface function;
    values are chosen at random, while the above table indicates which properties where shared across which robot systems.
    {We} refer to Appendix~\ref{app:robots-rotational} for details on the 2- and 3-link rotational robots' dynamics
    and to Appendix~\ref{app:robots-prismatic} for details on the 2-link prismatic robots' dynamics.
    }
    \label{tab:robots}
\end{table}

\paragraph{Evaluation}
For each system $(F^i, \theta^i, A^i),\ i\in\{1, \ldots, 24\}$
and repetition $k\in\{1,\ldots,10\}$,
running the system using the participants' controller
and comparing the realized end-effector trajectory against a target process $z_*^{i,k}:[0,2]\rightarrow \mathbb{R}^2$
the following loss is computed\textsuperscript{\ref{sharedfootnote}}
\begin{equation}
    \label{eq:full_score_track2}
    J_{i,k}:=b_{i,k}\cdot {\int_{0}^{2}||Z^{i,k}(t)-z^{i,k}_*(t)||^2_2 dt } + c_{i,k}\cdot {\int_{0}^{2}U^{i,k}(t)^{\top}U^{i,k}(t) dt },
\end{equation}
where $b_{i,k}$ and $c_{i,k}$ are scaling constants which are selected such that $J_{i,k}=100$ when no controls are applied and $J_{i,k}=1$ if an oracle LQR-controller is used, that is, an LQR-controller using the true robot dynamics and interface function. If $J_{i,k}$ is smaller than $1$, it is improving on the oracle LQR-controller; if it is larger than $100$ the performance is worse than when doing nothing. We clip all scores at $100$ before averaging them. The scaling is meant to ensure that losses are comparable across each repetition.

For the (preliminary) leaderboard, which was updated during the competition, only $12$ systems were evaluated, and the mean loss across those systems (and all corresponding repetitions) was shown on the leaderboard. For the final ranking, the average loss across the $12$ held-out systems (and all corresponding repetitions) was used.

\subsection{Rotational robots}\label{app:robots-rotational}

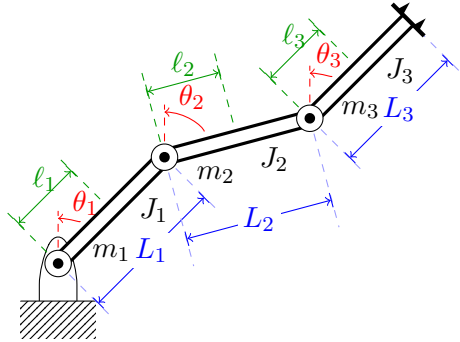
\begin{figure}
    \centering
    \input{tikz/3link_rot}
    \caption{Diagram of a 3-link rotational robotic arm.}
    \label{fig:robot_3link_arm}
\end{figure}

We consider two types of rotational robotic manipulators: open chain planar manipulators with three (cf.\ Figure~\ref{fig:robot_3link_arm}) and two revolute joints.
Joints can be controlled by applying a voltage signal to a DC motor located in the joint, which creates a torque.

We begin with discussing the 3-link manipulator and then show the simplified version for the 2-link variant.
Let $Z(t) = [ \theta_1(t), \theta_2(t), \theta_3(t), \omega_1(t), \omega_2(t), \omega_3(t) ]\transpose \in \mathbb{R}^6$ be the state of the robotic arm, consisting of joint angles ($\theta_1,\theta_2,\theta_3$) and corresponding angular velocities ($\omega_1,\omega_2,\omega_3$).
Let $U(t) = [ \tau_1(t), \tau_2(t), \tau_3(t) ]\transpose \in \mathbb{R}^3$ be the input joint torques ($\tau_1,\tau_2,\tau_3$).

The robotic arm is characterized by the following properties for the links, $i\in\{1,..,3\}$:
\begin{itemize}
    \item $m_i$ is the link mass,
    \item $J_i$ is the link rotational moment of inertia,
    \item $L_i$ is the link length,
    \item $\ell_i$ is the location of the link center of mass, and
    \item $c_i$ is the joint rotational friction coefficient.
\end{itemize}

The second-order dynamic system of the robotic arm is expressed through the following set of first-order equations \citep{jian2003dynamic}:
\begin{align*}
    \tfrac{\text{d}}{\text{dt}}[\theta_1] = \omega_1\\
    \tfrac{\text{d}}{\text{dt}}[\theta_2] = \omega_2\\
    \tfrac{\text{d}}{\text{dt}}[\theta_3] = \omega_3\\
    \tfrac{\text{d}}{\text{dt}}[\omega_1] = \alpha_1\\
    \tfrac{\text{d}}{\text{dt}}[\omega_2] = \alpha_2\\
    \tfrac{\text{d}}{\text{dt}}[\omega_3] = \alpha_3
\end{align*}
The joint acceleration terms $\alpha = [ \alpha_1, \alpha_2, \alpha_3 ]\transpose$ are determined via:
\begin{align}
    \label{eq:acc_rot}
    \alpha = M^{-1}\left( \tau - C \omega - N \right)
\end{align}
where the inertia matrix $M$, Coriolis matrix $C$, and external force vector $N$ are:
\begin{gather*}
    M = \begin{bmatrix}
        M_{11} & M_{12} \cos(\theta_2 - \theta_1)  & M_{13} \cos(\theta_3 - \theta_1) \\
        M_{12} \cos(\theta_2 - \theta_1) & M_{22} & M_{23} \cos(\theta_3 - \theta_2) \\
        M_{13} \cos(\theta_3 - \theta_1) & M_{23} \cos(\theta_3 - \theta_2) & M_{33} \end{bmatrix}, \\
    C = \begin{bmatrix}
        0 & C_{12} \sin(\theta_2 - \theta_1) \omega_2 & C_{13} \sin(\theta_3 - \theta_1) \omega_3 \\
        C_{21} \sin(\theta_2 - \theta_1) \omega_1 & 0 & C_{23} \sin(\theta_3 - \theta_2) \omega_3 \\
        -C_{13} \sin(\theta_3 - \theta_1) \omega_1 & C_{32} \sin(\theta_3 - \theta_2) \omega_2 & 0 \end{bmatrix},\text{ and} \\
    N = \begin{bmatrix}
        N_1 \sin(\theta_1) + c_1 \omega_1 \\
        N_2 \sin(\theta_2) + c_2 \omega_2 \\
        N_3 \sin(\theta_3) + c_3 \omega_3
        \end{bmatrix}
\end{gather*}
with coefficients
\begin{align*}
    M_{11} &= m_1 \ell_1^2 + J_1 + (m_2 + m_3) L_1^2 \\
    M_{12} &= (m_2 \ell_2 + m_3 L_2) L_1 \\
    M_{13} &= m_3 \ell_3 L_1 \\
    M_{22} &= m_2 \ell_2^2 + J_2 + m_3 L_2^2 \\
    M_{23} &= m_3 \ell_3 L_2 \\
    M_{33} &= m_3 \ell_3^2 + J_3 \\
    C_{12} &= - (m_2 \ell_2 + m_3 L_2) L_1 \\
    C_{13} &= - m_3 \ell_3 L_1 \\
    C_{21} &= (m_2 \ell_2 + m_3 L_2) L_1 \\
    C_{23} &= - m_3 \ell_3 L_2 \\
    C_{32} &= m_3 \ell_3 L_2 \\
    N_{1} &= - (m_1 \ell_1 + (m_2 + m_3) L_1) g\\
    N_{2} &= - (m_2 \ell_2 + m_3 L_2) g\\
    N_{3} &= - m_3 \ell_3 g
\end{align*}
and $g$ is gravitational acceleration.

For the 2-link robot, we omit all terms that correspond to the third joint.
That is, the acceleration $\alpha = [ \alpha_1, \alpha_2 ]\transpose$ is still given by~\eqref{eq:acc_rot}, but we now need to adapt $M$, $C$, and $N$.
When only considering two joints, we have
\begin{align*}
    M &= \begin{bmatrix}
    M_{11} & M_{12}\cos(\theta_2 - \theta_1)\\
    M_{12} \cos(\theta_2 - \theta_1) & M_{22}
    \end{bmatrix}\\
    C&= \begin{bmatrix}
    0 & C_{12}\sin(\theta_2 - \theta_1)\omega_2\\
    C_{21}\sin(\theta_2-\theta_1)\omega_1 & 0
    \end{bmatrix}\\
    N &= \begin{bmatrix}
    N_1\sin(\theta_1) + c_1\omega_1\\
    N_2\sin(\theta_2) + c_2\omega_2
    \end{bmatrix}
\end{align*}
with coefficients
\begin{align*}
    M_{11} &= m_1\ell_1^2 + J_1 + m_2L_1^2\\
    M_{12} &= m_2\ell_2L_1\\
    M_{22} &= m_2\ell_2^2 + J_2\\
    C_{12} &= -m_2\ell_2^2L_1\\
    C_{21} &= m2\ell_2L_1\\
    N_1 &= -(m_1\ell_1 + m_2L_1)g\\
    N_2 &= -m_2\ell_2g.
\end{align*}

\begin{remark}
Note that in the open source code, we define the joint angles of the 2-link manipulator with respect to each other instead of with respect to the vertical axis (\cite{murray2017mathematical}).
Hence, the equations slightly differ from the ones presented above.
However, the dynamics are the same as described herein.
\end{remark}

\subsection{Prismatic robot}\label{app:robots-prismatic}

\begin{figure}
    \centering
    \input{tikz/2link_pris}
    \caption{Diagram of the 2-link prismatic robot arm.}
    \label{fig:robot_2link_pris}
\end{figure}
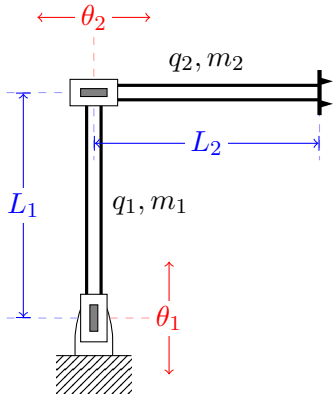

Besides the two versions of rotational robots, we also consider a 2-link prismatic robot arm (Figure~\ref{fig:robot_2link_pris}). This idealized prismatic robot is actuated by prismatic joints that change the link lengths, such that $L_i = q_i + \theta_i$, where $q_i$ represents the link length at zero joint input. Although the link length changes, we assume the link mass $m_i$ remains constant.

Due to the lateral instead of rotational movements, the dynamics of this robot are considerably simpler.
The joint acceleration terms $\alpha = [ \alpha_1, \alpha_2 ]\transpose$ are given by \begin{align}
    \alpha = M^{-1}\left( \tau - N \right).
\end{align}
The equation is similar to~\eqref{eq:acc_rot}, but without the Coriolis term since there are only lateral movements.
Mass matrix and external force vector are
\(
M = \begin{bmatrix}
m_1 + m_2 & 0\\
0 & m_2
\end{bmatrix}
\)
and
\(
N = \begin{bmatrix}
g(m_1 + m_2)\\ 0
\end{bmatrix},
\)
respectively.

\subsection{ROBO results}
\label{ssec:roboresults}

The results from Track ROBO are summarized below. The keywords were chosen by the organizers based on the participants' descriptions of their approaches. The score function is standardized such that a value of $1$ corresponds to the performance of an oracle LQR-controller
using the true system (optimizing only the trajectory, not the cost). A score of 100 corresponds to the zero solution, $U \equiv 0$.

\begin{center}
\begin{tabular}{ c |  c | c | c }
Team name & Score & Place & Keywords \\ \hline
  Ajoo  & 0.918 & 1st &  Estimation of robot dynamics  \\ \hline
  TeamQ  & 16.121 & 2nd & Neural network prediction \\  \hline
 jmunozb  & 29.539 &  3rd & \begin{tabular}{c}  Linear system approximation \\ Regression with polynomial features \end{tabular}
\end{tabular}
\end{center}
\end{document}

%% file: tikz/3link_rot.tex
\newcommand{\nvar}[2]{%
    \newlength{#1}
    \setlength{#1}{#2}
}

\nvar{\dg}{0.3cm}
\def\dw{0.25}\def\dh{0.5}
\nvar{\ddx}{1cm}

\def\link{\draw [double distance=1.5mm, very thick] (0,0)--}
\def\joint{%
    \filldraw [fill=white] (0,0) circle (5pt);
    \fill[black] circle (2pt);
}
\def\grip{%
    \draw[ultra thick](0cm,\dg)--(0cm,-\dg);
    \fill (0cm, 0.5\dg)+(0cm,1.5pt) -- +(0.6\dg,0cm) -- +(0pt,-1.5pt);
    \fill (0cm, -0.5\dg)+(0cm,1.5pt) -- +(0.6\dg,0cm) -- +(0pt,-1.5pt);
}
\def\robotbase{%
    \draw[rounded corners=8pt] (-\dw,-\dh)-- (-\dw, 0) --
        (0,\dh)--(\dw,0)--(\dw,-\dh);
    \draw (-0.5,-\dh)-- (0.5,-\dh);
    \fill[pattern=north east lines] (-0.5,-1) rectangle (0.5,-\dh);
}

\newcommand{\angann}[4]{%
    \begin{scope}[red]
    \draw [dashed, red, rotate=#1] (0,0) -- (0pt,0.7\ddx);
    \draw [<-, shorten >=5.5pt, rotate=#1] (0pt,0.6\ddx) arc (90:-#3:0.6\ddx);
    \node at (-#4/2-2:0.6\ddx+8pt) {#2};
    \end{scope}
}

\newcommand{\lineann}[4][0.5]{%
    \begin{scope}[rotate=#2, blue,inner sep=2pt]
        \draw[dashed, blue!40] (0,0) -- +(0,#1)
            node [coordinate, near end] (a) {};
        \draw[dashed, blue!40] (#3,0) -- +(0,#1)
            node [coordinate, near end] (b) {};
        \draw[|<->|] (a) -- node[fill=white] {#4} (b);
    \end{scope}
}

\newcommand{\lineannshort}[4][0.5]{%
    \begin{scope}[rotate=#2, green!60!black,inner sep=2pt]
        \draw[dashed, green!60!black] (0,0) -- +(0,#1)
            node [coordinate, near end] (a) {};
        \draw[dashed, green!60!black] (0.5*#3,0) -- +(0,#1)
            node [coordinate, near end] (b) {};
        \draw[|<->|] (a) -- node[above] {#4} (b);
    \end{scope}
}

\newcommand{\nolineann}[5][0.5]{%
    \begin{scope}[rotate=#2, inner sep=2pt]
        \node at(0.3*#3, #1){#4};
        \node at(0.7*#3, #1){#5};
    \end{scope}
}

\def\thetaone{45}
\def\Lone{2}
\def\thetatwo{-30}
\def\Ltwo{2}
\def\thetathree{30}
\def\Lthree{1.85}

\begin{tikzpicture}
    \robotbase
    \angann{0}{$\theta_1$}{-\thetaone}{-3*\thetaone}
    \lineann[-1]{\thetaone}{\Lone}{$L_1$}
    \nolineann[-0.4]{\thetaone}{\Lone}{$m_1$}{$J_1$}
    \lineannshort[1]{\thetaone}{\Lone}{$\ell_1$}
    \link(\thetaone:\Lone);
    \joint
    \begin{scope}[shift=(\thetaone:\Lone), rotate=\thetaone]
        \angann{-\thetaone}{$\theta_2$}{-30-\thetaone-\thetatwo}{-\thetaone}
        \lineann[-1.5]{\thetatwo}{\Ltwo}{$L_2$}
        \nolineann[-0.4]{\thetatwo}{\Ltwo}{$m_2$}{$J_2$}
        \lineannshort[1.2]{\thetatwo}{\Ltwo}{$\ell_2$}
        \link(\thetatwo:\Ltwo);
        \joint
        \begin{scope}[shift=(\thetatwo:\Ltwo), rotate=\thetatwo]
            \angann{-\thetaone-\thetatwo}{$\theta_3$}{-\thetaone-\thetatwo-\thetathree}{-\thetaone-\thetatwo}
            \lineann[-1]{\thetathree}{\Lthree}{$L_3$}
            \nolineann[-0.4]{\thetathree}{\Lthree}{$m_3$}{$J_3$}
            \lineannshort[1]{\thetathree}{\Lthree}{$\ell_3$}
            \link(\thetathree:\Lthree);
            \joint
            \begin{scope}[shift=(\thetathree:\Lthree), rotate=\thetathree]
                \grip
            \end{scope}
        \end{scope}
    \end{scope}
\end{tikzpicture}

%% file: tikz/2link_pris.tex
\newcommand{\nvar}[2]{%
    \newlength{#1}
    \setlength{#1}{#2}
}

\nvar{\pdg}{0.3cm}
\def\dw{0.25}\def\dh{0.5}
\def\link{\draw [double distance=1.5mm, very thick] (0,0)--}
\def\joint{%
    \filldraw [fill=white] (-5pt,-9pt) rectangle ++(10pt,18pt);
    \filldraw [fill=gray] (-1.5pt,-5pt) rectangle ++(3pt,10pt);
}
\def\grip{%
    \draw[ultra thick](0cm,\pdg)--(0cm,-\pdg);
    \fill (0cm, 0.5\pdg)+(0cm,1.5pt) -- +(0.6\pdg,0cm) -- +(0pt,-1.5pt);
    \fill (0cm, -0.5\pdg)+(0cm,1.5pt) -- +(0.6\pdg,0cm) -- +(0pt,-1.5pt);
}
\def\robotbase{%
    \draw[rounded corners=8pt] (-\dw,-\dh)-- (-\dw, 0) --
        (0,\dh)--(\dw,0)--(\dw,-\dh);
    \draw (-0.5,-\dh)-- (0.5,-\dh);
    \fill[pattern=north east lines] (-0.5,-1) rectangle (0.5,-\dh);
}

\newcommand{\angannpris}[4][0.5]{%
    \begin{scope}[rotate=#2, red, inner sep=2pt]
        \draw[dashed, red!40] (0,0) -- +(0,#1);
        \draw[<->] (-#3/4, #1) -- node[fill=white] {#4} (#3/4, #1);
    \end{scope}
}

\newcommand{\lineann}[4][0.5]{%
    \begin{scope}[rotate=#2, blue,inner sep=2pt]
        \draw[dashed, blue!40] (0,0) -- +(0,#1)
            node [coordinate, near end] (a) {};
        \draw[dashed, blue!40] (#3,0) -- +(0,#1)
            node [coordinate, near end] (b) {};
        \draw[|<->|] (a) -- node[fill=white] {#4} (b);
    \end{scope}
}

\newcommand{\lineannshort}[4][0.5]{%
    \begin{scope}[rotate=#2, green!60!black,inner sep=2pt]
        \draw[dashed, green!60!black] (0,0) -- +(0,#1)
            node [coordinate, near end] (a) {};
        \draw[dashed, green!60!black] (0.5*#3,0) -- +(0,#1)
            node [coordinate, near end] (b) {};
        \draw[|<->|] (a) -- node[above] {#4} (b);
    \end{scope}
}

\newcommand{\textann}[4][0.5]{%
    \begin{scope}[rotate=#2, inner sep=2pt]
        \draw (0.5*#3, #1) node {#4};
    \end{scope}
}

\def\thetaone{90} %
\def\Lone{3}
\def\thetatwo{-90}
\def\Ltwo{3}

\begin{tikzpicture}
    \robotbase
    \angannpris[-1.]{\thetaone}{\Lone}{$\theta_1$}
    \lineann[1.25]{\thetaone}{\Lone}{$L_1$}
    \link(\thetaone:\Lone);
    \textann[-0.75]{\thetaone}{\Lone}{$q_1, m_1$}
    \joint
    \begin{scope}[shift=(\thetaone:\Lone), rotate=\thetaone]
        \angannpris[1.]{\thetatwo}{\Ltwo}{$\theta_2$}
        \lineann[-0.9]{\thetatwo}{\Ltwo}{$L_2$}
        \link(\thetatwo:\Ltwo);
        \textann[0.4]{\thetatwo}{\Ltwo}{$q_2, m_2$}
        \joint
        \begin{scope}[shift=(\thetatwo:\Ltwo), rotate=\thetatwo]
            \grip
        \end{scope}
    \end{scope}
\end{tikzpicture}